\documentclass[conference]{IEEEtran}
\IEEEoverridecommandlockouts
\usepackage{amsmath,amssymb,amsfonts}
\usepackage{algorithmic}
\usepackage{algorithm}
\usepackage{graphicx}
\usepackage{textcomp}
\usepackage{xcolor}
\usepackage{amsthm}
\usepackage{booktabs}
\usepackage{multirow}
\usepackage{pifont}
\usepackage{amssymb}
\usepackage{float}
\newtheorem{theorem}{Theorem}
\newtheorem{assumption}{Assumption}
\newtheorem{lemma}{Lemma}
\newtheorem{remark}{Remark}
\usepackage[hyphens]{url}

\usepackage[square,numbers,sort&compress]{natbib}

\newcommand{\normtwo}[1]{\left \lVert #1 \right \rVert}
\newcommand{\vecdot}[2]{\langle #1,#2 \rangle}
\newcommand{\cmark}{\ding{51}}%
\newcommand{\xmark}{\ding{55}}%

\def\BibTeX{{\rm B\kern-.05em{\sc i\kern-.025em b}\kern-.08em
    T\kern-.1667em\lower.7ex\hbox{E}\kern-.125emX}}
\begin{document}

\title{Lightweight and Robust Federated Data Valuation}

\author{\IEEEauthorblockN{Guojun Tang}
\IEEEauthorblockA{\textit{University of Calgary} \\
Calgary, Canada \\
guojun.tang@ucalgary.ca}
\and
\IEEEauthorblockN{Jiayu Zhou}
\IEEEauthorblockA{\textit{University of Michigan} \\
Ann Arbor, United States \\
jiayuz@umich.edu}
\and
\IEEEauthorblockN{Mohammad Mamun}
\IEEEauthorblockA{\textit{National Research Council Canada	} \\
Fredericton, Canada \\
mohammad.mamun@nrc-cnrc.gc.ca}
\and
\IEEEauthorblockN{Steve Drew}
\IEEEauthorblockA{\textit{University of Calgary} \\
Calgary, Canada \\
steve.drew@ucalgary.ca	}
}

\maketitle

\begin{abstract}
Federated learning (FL) faces persistent robustness challenges due to non-IID data distributions and adversarial client behavior. A promising mitigation strategy is contribution evaluation, which enables adaptive aggregation by quantifying each client’s utility to the global model. However, state-of-the-art Shapley-value-based approaches incur high computational overhead due to repeated model reweighting and inference, which limits their scalability. We propose FedIF, a novel FL aggregation framework that leverages trajectory-based influence estimation to efficiently compute client contributions. FedIF adapts decentralized FL by introducing normalized and smoothed influence scores computed from lightweight gradient operations on client updates and a public validation set. Theoretical analysis demonstrates that FedIF yields a tighter bound on one-step global loss change under noisy conditions. Extensive experiments on CIFAR-10 and Fashion-MNIST show that FedIF achieves robustness comparable to or exceeding SV-based methods in the presence of label noise, gradient noise, and adversarial samples, while reducing aggregation overhead by up to 450x. Ablation studies confirm the effectiveness of FedIF’s design choices, including local weight normalization and influence smoothing. Our results establish FedIF as a practical, theoretically grounded, and scalable alternative to Shapley-value-based approaches for efficient and robust FL in real-world deployments.
\end{abstract}


\section{Introduction}


FL \cite{fedavg} has achieved great success in a wide range of applications from edge computing, Internet of Things (IoT), to healthcare and finance \cite{cross_devices_1, cross_devices_2,cross_silos_1,cross_silos_2}. Among these application areas, practical challenges remain. Unlike the traditional centralized learning framework, FL participants commonly suffer from non-independent and identically distributed (non-IID) data distributions \cite{non_iid}, resulting in higher convergence times and lower accuracy. Moreover, some clients may not act fully trusted and may even be malicious while participating in the collaborative training. Common vulnerable conditions are associated with noisy clients \cite{noisy_label} and gradient attacks \cite{adv_fl_survey}, which degrade the model's performance and pose a threat to the robustness of collaborative training systems. It is essential to consider vulnerable conditions in the non-IID setting \cite{fednoisy,fednoro} in FL, with the aim of providing robustness in heterogeneous data and vulnerable environments.


Addressing the issue of statistically non-IID data across FL clients has been well-discussed. Existing works, such as FedProx \cite{fedprox}, SCAFFOLD \cite{scaffold}, and MOON \cite{moon}, have demonstrated their effectiveness in mitigating the impact of data heterogeneity.
We note that these algorithms assume that participants are honest and do not engage in adversarial behavior. Shapley Value (SV) \cite{shapley_value}, a method to quantify the contribution of each participant in cooperative game theory, is widely applied to the evaluation of client contribution \cite{data_shapley,fedsv}.
In FL, SV has been shown to be practical in detecting noisy clients and adversarial participants \cite{fedsv}. Existing works have proposed SV-based adaptive weight model averaging algorithms to provide a more robust aggregation \cite{AFedSV,localFedSV,ShapFed}.
One caveat impeding the practical use of these algorithms is that they primarily rely on the time-consuming reweighting-based approach.
During each global training epoch, the server repeatedly reweights the global model parameters and evaluates the model's performance (e.g., classification accuracy) using a validation set.
Through this iterative reweighting, the server can assess the individual contributions of clients to the global model. 
However, model reweighting and repeated model inferences are time-consuming and inefficient, limiting both scalability and efficiency in practical deployments.

\textit{Does a more efficient method exist for data valuation in FL?} Recently, \citet{tracIn} proposed an alternative data valuation method, \textit{TracIn}. This proposed method identifies the influence of training data on a model by the first-order gradient approximation. Compared to SV in FL, this gradient-based method is more efficient because it requires only the simple calculation of gradients, rather than repeated model reweighting and inference. It has demonstrated feasibility in various tasks, such as detecting mislabeled data and selecting high-quality data \cite{less}. 

Inspired by TracIn, we introduce a gradient-based data valuation method to FL and design an effective and robust FL aggregation strategy based on it.
In this paper, we propose FedIF, an effective and robust model aggregation algorithm for FL based on the trajectory data \textbf{I}n\textbf{F}luence. Specifically, we first analyze the existing data valuation method TracIn, which estimates a single data point by accumulating the trajectory of the approximated loss function from different checkpoints. Distinct from TracIn for a single data point, we focus on evaluating client updates during the FL protocol. Specifically, we utilize the idea of trajectory influence to calculate the influence score of each participant by verifying the local update with a validation gradient. In addition, we employ L2 normalization for each updated model, which helps us reveal fine-grained differences when estimating the local update. Client selection in FL usually varies across different global epochs in FL. To guarantee fair comparison among clients from different epochs, we characterize our proposed data valuation method by applying normalization of the influence score and a smooth update \cite{AFedSV}. Once we have the influence score, we can quantify the contribution of each client to the system's overall performance. Therefore, we can dynamically calculate the adaptive averaging weights to aggregate local models, ensuring model robustness. The client identified as a significant contributor merits more weight during aggregation, while clients with smaller contributions will be considered less in the model aggregation. The calculation of the influence score is efficient and only requires a few simple operations on the updated weight and the validation gradient. Through extensive empirical studies, we find that FedIF achieves comparable or better performance compared with other FL data valuation methods \cite{fedsv,AFedSV} with substantially smaller computational overhead. Our contributions are summarized as follows.
\begin{itemize}
    \item We adopt a lightweight data valuation method based on trajectory influence for FL by applying normalization and smooth updates, enabling it to evaluate local updates from clients. To our knowledge, this is the first work to utilize trajectory influence in estimating federated influence scores.
in aggregation time
    \item We design an adaptive averaging weight aggregation rule based on the proposed federated data valuation method. The theoretical analysis shows that this adaptive averaging weight-based method has a tighter upper bound on the one-step loss change of the global model compared to FedAvg in noisy environments. Therefore, it preserves its robustness.
    \item Experimental results demonstrate the outstanding efficiency of FedIF, being up to 450x more efficient compared to SV-based data valuation methods. Meanwhile, FedIF achieves robustness and efficiency in most vulnerable settings, such as those with label noise and gradient noise. Ablation study also indicates the efficacy of each module in our proposed work.
\end{itemize}

The rest of this paper is structured as follows. We first discuss the related work of federated learning and data valuation in Section \ref{literature_review}, followed by an overview of the preliminaries of federated learning and trajectory data influence in Section \ref{preliminaries}. Section \ref{methodology} is dedicated to the presentation of the FedIF methodology along with a discussion of its theoretical analysis. In Section \ref{experiment}, we undertake empirical experimentation with our algorithm utilizing image classification datasets. Concluding the paper, we summarize the findings of FedIF and explore future works.

\begin{figure*}[th]
    \centering
    \includegraphics[scale=0.5,trim={20 60 5 60},clip]{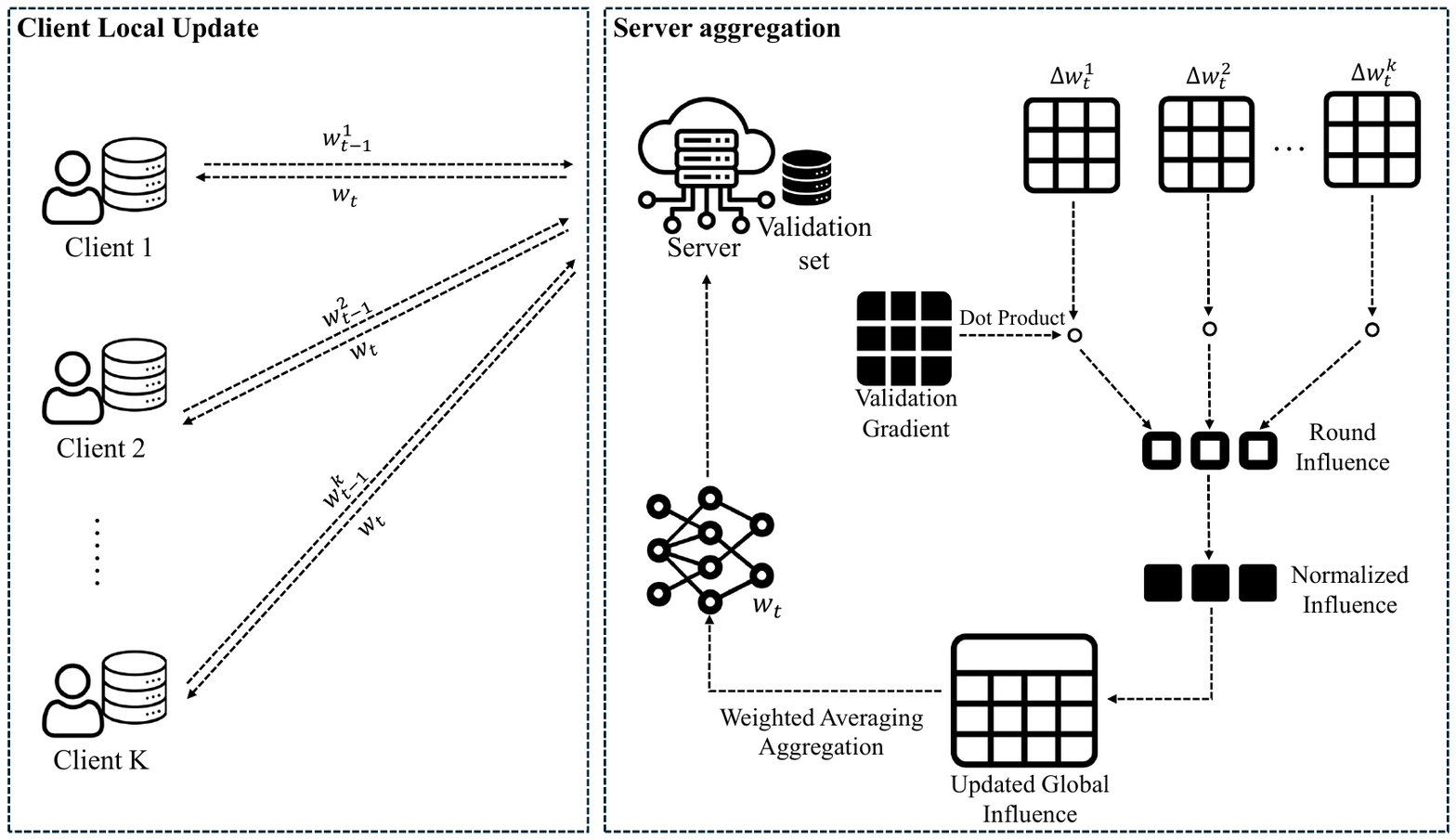}
    \caption{Overview of FedIF, a novel FL aggregation framework that leverages trajectory-based influence estimation to efficiently compute client contributions.}
    \label{overview}
\end{figure*}

\section{Related Work}\label{literature_review}
FL has garnered significant attention because it enables collaborative training among multiple clients while avoiding the sharing of raw data, thereby preserving privacy and enhancing communication efficiency. It still leaves a series of open challenges, such as data heterogeneity \cite{non_iid, fedprox,scaffold,moon}, privacy \cite{privacy_survey, dlg, batchcrypt}, and robustness \cite{fednoro,AFedSV,fltrust,krum,noisy_label}. In our paper, we focus on the robustness of ML, which provides a robust training method for various adversarial settings.

In FL, the malicious client poses a threat to collaborative training through various attacks, for example, data poisoning attacks by flipping the label of training data \cite{fednoro, noisy_label} or training with adversarial samples \cite{pgd_fl}, and model poisoning by uploading corrupted models \cite{model_poisoning}. Those attacks usually lead to the downgrade of model performance, decreasing the accuracy of model prediction, decelerating, or even failing model convergence. Some existing work in FL aims to alleviate the negative impact of these attacks. For example, Wu et al. \cite{fednoro} designed FedNoRo, a robust FL algorithm based on label noise that uses a two-stage framework to mitigate the impact of noisy clients. Cao et al. \cite{fltrust} proposed a Byzantine-robust FL method called FLTrust that leveraged a validation set to calculate the cosine similarity with local updates from clients. The model aggregation would be calculated by the weighted averaging of the local update based on similarity values. In addition to the aforementioned schemes, the data valuation methods are also effective at detecting the data quality of clients, assisting us to improve the robustness of FL.

Common data valuation methods are as follows: Shapley value \cite{data_shapley}, influence function \cite{influence_function,blackbox_influence}, and trajectory data influence \cite{tracIn}. The influence function estimates how much removing or perturbing a single training point affects the loss on a test data point and usually requires the inverse of the Hessian matrix from the training data. It has been widely used in the centralized ML scenario. If we adopt it in FL, it requires an additional calculation for the Hessian matrix from the client, which is hard to verify. Therefore, this method is not suitable to apply directly to the FL. Shapley value is a classic concept in cooperative game theory to compute the average marginal contribution of a player across all possible subsets of participants in this game. It has been widely used to measure the clients' contribution to the global model in FL. The Shapley value demonstrates significant efficacy in evaluating the contribution of clients. Intuitively, clients with bigger contributions merit higher averaging weights during the model aggregation. Existing works \cite{AFedSV,gtg_shapley,ShapFed,FAIR} developed adaptive weight model aggregation to improve FL performance based on this idea. To compute the ground-truth Shapley value in FL, it requires aggregating the model parameters across all possible subsets of participants and performing model inference across the model across all subsets, which is computationally intensive. Even though we may approximate the Shapley value by applying the Monte Carlo algorithm \cite{AFedSV,fedsv} or evaluating by partial gradients \cite{gtg_shapley,ShapFed}, the execution time and computational overhead are still unignorable. Pruthi et al. \cite{tracIn} proposed TracIn, a lightweight data valuation method by tracing the dot product of training data gradients and test data gradients of each checkpoint model. This method was not only effective in data valuation but also computationally efficient. Compared to the Shapley value, it only simply calculates the dot product of gradients instead of repeatedly reweighting and inferring the model.

\section{Preliminaries}\label{preliminaries}
\subsection{Federated Learning}
Most FL frameworks adopt the standard aggregation method called FedAvg \cite{fedavg}. In this context, the system comprises a set of local clients and a central coordinator server that collaboratively trains a global model without sharing the raw data. We assume that there are $K$ clients in the system and that each client $i$ is equipped with a local training dataset $D^i$, where $i=1,\dots, K$. Under the coordination of the central server, FL aims to achieve the global training objective $F(w)$ that minimizes the local training objective $F^i(w)$ of the client $i$ as follows: 
$$\underset{w \in \mathbb{R}^d}{arg\,min}\ F(w)=\sum^{K}_{i=0}p^iF^i(w), $$
where $w$ denotes the model parameters and $p^k$ is the averaging weight of the client $i$. We adopt the training loss $F^i(w)=\ell(D^i;w)$ as the local objective of each client. In FedAvg, clients and the server apply the following protocol to train the global model: 
\begin{enumerate}
    \item Each client first initializes by downloading the initial model parameters from the central server. 
    \item A small portion of clients, randomly selected by the central server, use their local dataset to train a local model.
    \item The central server collects those local model parameters, aggregates them by averaging them into one global model, and dispatches them to all clients. 
\end{enumerate}
These interactions between clients and the central server repeat for multiple rounds until a specific global training epoch is reached.

\subsection{Trajectory Data Influence}
The approach from \citet{tracIn} utilizes the first-order gradient approximation on test data to estimate the influence of training data. Given a test example $d'$, the change of the loss in the epoch $t$ can be estimated by the first-order Taylor approximation as follows:
\begin{equation}\label{changeloss}
    \ell(w_{t+1},d') \approx \ell(w_{t},d') + \nabla \ell(w_t,d') \cdot (w_{t+1}-w_t).
\end{equation}

When we use the optimizer to update the model, the change in the model parameters can be denoted as $w_{t+1}-w_t= - \eta_t \nabla \ell(w_t,d_t)$, where $d_t$ is the training data point and $\eta_t$ is the learning step at epoch $t$. After substitution and rearrangement, we denote the change in the loss by
\begin{equation}
    \ell(w_{t},d') - \ell(w_{t+1},d') \approx  \eta_t \nabla \ell(w_t,d_t)  \cdot \nabla \ell(w_t,d').
\end{equation}
The fundamental theorem of calculus reveals that the integral of a function's gradients between two points is equivalent to the difference in the function's values at those points \cite{tracIn}. Therefore, the cumulative influence of training examples on a test data point is equal to the total reduction in loss at this test point during the training process. Since the training data point $d_t$ is used to update the parameter, we can identify the approximate influence $d_t$ on the final model by summing up these loss changes at each epoch. Commonly, we calculate this influence by the checkpoint models: assume that we have checkpoint models from epoch $t=1,2,...,T$. Under the test data point $z'$, the influence of the training data point $d_t$ on the model can be represented by 
\begin{equation}\label{tracincp}
    INF(d,d')=\sum_{t=1}^T\eta_t \nabla \ell(w_t,d) \cdot \nabla \ell(w_t,d')
\end{equation}

\section{Methodology}\label{methodology}

In this section, we present our robust FL framework based on data influence. We first propose a method to evaluate how local updates from clients influence the global model using the local model parameters and the gradient from the validation set. Then, the server adjusts the averaging weight adaptively according to these influence values to provide a more robust model aggregation. Finally, we provide the theoretical analysis and explain how our method enhances the robustness of FL. The overview illustration is shown in Figure \ref{overview}.

\subsection{Local Data Influence in FL}

Recall Equation (\ref{tracincp}), we estimate the influence of the local dataset $D^i$ from the client $i$ by the validation set $D'$ to the global model: 
\begin{equation}\label{inf_1}
    INF_i=Inf(D^i,D')=\sum_{t=1}^T\eta_t \nabla \ell(w_t,D^i) \cdot \nabla \ell(w_t,D'),
\end{equation}
where the global model in each training epoch $w_t$ is treated as the checkpoint for the measurement. It is noted that clients upload their updated model parameters to the server instead of their gradients in the typical FL framework. We denote the updated model parameters of the client $i$ at epoch $t$ as $w^i_{t}= w_{t-1}- \eta_{t-1} \nabla \ell(w_{t-1},D^i)$. Substitute these parameters into Equation (\ref{inf_1}) and we have the \textbf{accumulative influence} of $D^i$ to the global model: 
\begin{equation}\label{acc_inf}
    INF_i=\sum_{t=1}^T (w_{t-1}-w^i_{t}) \cdot \nabla \ell(w_{t-1},D').
\end{equation}
The accumulative influence measures the influence of the local data from a client in an idealized situation. However, clients are typically sampled randomly by the server to participate in the training in FL, and the influence values vary across training epochs. Based on the idealized accumulative influence, we propose a practical method for evaluating the influence of each client's local data, which is adaptive to the FL. 

First, we define the \textbf{round influence $\Phi$} of the $D^i$ to the global model $w_t$ at the training round $t$ as follows:
\begin{equation}\label{rinf}
    \Phi^i_t= \frac{\Delta w_t^i}{\normtwo{\Delta w_t^i}} \cdot \nabla \ell(w_{t-1},D'),
\end{equation}
where $\Delta w_t^i=w_{t-1}-w^i_{t}$. The round influence is the dot product of the updated local model and the validation gradient. It evaluates how the local data influences the change of the global model at this round. The local update may vary widely in scale at a global training epoch. Therefore, we utilize the L2 norm to normalize the magnitude of the local update, defined as the local \textbf{weight normalization}. 

When we calculate the round-wise influence for the current participants, value ranges are commonly varied by different global epochs. We also need to ensure a fair comparison of the influence values from different rounds. Subsequently, we apply min-max normalization to map the values into a specific range. We define the \textbf{normalized round influence $\Psi$} in each global epoch \cite{AFedSV, fedsv} as follows:
\begin{equation}\label{ninf}
    \Psi^i_t= \frac{\Phi^i_t-min(\Phi_t)}{max(\Phi_t)-min(\Phi_t)},
\end{equation}
where $max(\Phi_t)$ and $min(\Phi_t)$ denote the maximum and minimum round influence among all participants at the global epoch $t$, respectively.

The influence value is treated as a reference for adjusting the averaging weights during model aggregation. The calculated values may be unstable and result in oscillatory averaging weights. As suggested by \cite{AFedSV,fedadp, ShapFed}, we adopt the smooth update strategy to stabilize the update of the influence value. Instead of directly accumulating these values, we use the smooth update strategy to calculate the final \textbf{global influence $\Omega$} as follows: 

\begin{equation}\label{ginf}
\Omega_t^i = 
\begin{cases} 
(1 - \gamma) \cdot \Omega_{t-1}^i + \gamma \cdot \Psi^i_t, & \text{if } i \in S_t, \\
\Omega_{t-1}^i, & \text{if } i \notin S_t, \\
0,  & \text{if } t=0.
\end{cases}
\end{equation}
In Equation (\ref{ginf}), $\gamma$ is the hyperparameter controlling the smooth update that determines how the influence value from the previous round will be considered at the current round's update. The values of the global influence of all clients are initialized as 0. If the client $i$ is selected to participate in training in the epoch $t$, denoted by $i \in S_t$, the server will first calculate its normalized influence value and then update its corresponding global influence by the smooth update strategy. Otherwise, the other clients who do not participate in the training of this epoch will maintain the same global influence values as the previous round. 

\subsection{Adaptive Weights Aggregation by Influence Scores}

The server can assess the contribution of each client according to Equations (\ref{rinf}), (\ref{ninf}), and (\ref{ginf}) by using the validation set. The client with a higher influence score can be considered as a more positive contribution to the global model and merits a higher averaging weight during the model aggregation. Therefore, we can determine the model aggregation weights for the global epoch $t$ based on influence scores:
\begin{equation}\label{avg_weights}
    p_t^i=\frac{\Omega^i_t}{\sum_{j=1}^{K}\Omega^j_{t}}.
\end{equation}
After computing the averaging weight, we aggregate the global model as $w_t =\sum_{i \in S_t} p_t^i w_t^i $. The entire procedure is presented in Algorithm \ref{fedif_alg}. Our proposed method follows a similar process to FedAvg \cite{fedavg}, including model initialization, random selection of the client, local update of the client, and iterative training. The significant modification is that our method calculates the adaptive averaging weights during the model aggregation according to the influence scores from clients.

\begin{algorithm}
\caption{The FedIF Algorithm}
\label{fedif_alg}

\textbf{Procedure: Federated Optimization} \\
\textbf{Input:} validation set $D'$, Clients $1...K$, local training epoch $E$, local training batch size $B$, local learning rate $\eta$, client fraction $C$, total global training epoch $T$, \\
\textbf{Output:} global model $w_T$

\begin{algorithmic}[1]
\STATE Server initializes global model $w_0$, and the $GINF$ as 0 for all clients
\FOR{each round $t = 1, 2, \ldots, T$}
    \STATE $m \leftarrow \max(C \times K, 1)$  
    \STATE $S_t \leftarrow$ Randomly select $m$ clients from $K$
    \FOR{each client $i \in S_t$ \textbf{in parallel}}
        \STATE $w_t^i \leftarrow \text{LocalTraining}(w_{t-1}, E, B, \eta)$ 
    \ENDFOR
    \FOR{each client $i \in S_t$}
        \STATE  $\Phi^i_t \leftarrow \frac{\Delta w_t^i}{\normtwo{\Delta w_t^i}} \cdot \nabla \ell(w_{t-1},D')$
    \ENDFOR
    \STATE $\Phi_t \leftarrow \{\Phi^i_t\}\ for\ i \in S_t$
    \STATE $\Psi^i_t \leftarrow \frac{\Phi^i_t-min(\Phi_t)}{max(\Phi_t)-min(\Phi_t)}$
    \STATE Update the $\Omega_t^i$ by Equation \ref{ginf} for all client 
    \STATE $p_t^i=\frac{\Omega^i_t}{\sum_{j=1}^{K}\Omega^j_{t}}$
    \STATE $w_t \leftarrow \sum_{i \in S_t} p^i w_t^i$ 
\ENDFOR
\STATE \textbf{return} Final global model $w^T$
\end{algorithmic}

\vspace{1em}
\textbf{Procedure: LocalTraining($w$, $E$, $B$, $\eta$)} \\
\textbf{Input:} Model $w$, local epochs $E$, local training batch size $B$, learning rate $\eta$ \\
\textbf{Output:} Updated local model $w^i$

\begin{algorithmic}[1]
\STATE Initialize local model weights $w_{\text{local}} \leftarrow w$
\FOR{each epoch $e = 1, \ldots, E$}
    \FOR{batch $(x, y)$ with size $B$ in local data}
        \STATE $g \leftarrow \nabla \ell(w_{\text{local}}; x, y)$ 
        \STATE $w_{\text{local}} \leftarrow w_{\text{local}} - \eta \cdot g$ 
    \ENDFOR
\ENDFOR
\STATE \textbf{return} $w_{\text{local}}$
\end{algorithmic}

\end{algorithm}

\subsection{Theoretical Analysis}\label{theoretical_analysis}

In this section, we present the theoretical analysis of FedIF, providing the upper bound of a one-step loss change for the global model and explaining how robustness is achieved compared to FedAvg. We first follow two common and typical assumptions in the theoretical analysis in FL \cite{fedavg_conver,fedprox,fedadp,rfa, localsgd}, listed below. 

\begin{assumption}[$L$-Lipschitz Smoothness]\label{l_smooth}
    The local objective $F^i(w)$ and the global objective $F(w)$ are of $L$-Lipschitz smoothness. For example, $\normtwo{\nabla F(w) -\nabla F(w')} \le L\normtwo{w-w'}$. 
\end{assumption}

\begin{assumption}[Bounded Local Disimilarity]\label{local_dissim}
    The local objective of a participant client $i$ has a bounded dissimilarity from the global objective: $\sum_ip^i\normtwo{\nabla F^i(w) - \nabla F(w)}^2 \le \beta^2$, where $\sum_ip^i=1$. 
\end{assumption}

\begin{lemma}\label{bounded_dissim}
    Using Assumption \ref{local_dissim}, we can obtain the other bounded relation between the local objective and the global objective $\sum_ip^i\normtwo{\nabla F^i(w) }^2 \le \beta^2 + \normtwo{\nabla F(w)}^2$.
\end{lemma}

The detailed proof of Lemma \ref{bounded_dissim} can be found in Appendix \ref{proof_bounded_dissi}.

To analyze the robustness of FL under noisy settings, we introduce an additional assumption that simplifies the local update, which may contain an additive noise term. 

\begin{assumption}[Noisy Client update]\label{client_noise}
    Client $i$ may return an update with noise $\delta^i_t$ at epoch $t$: $g_t^i=\nabla F^i(w) + \delta^i_t$. The noise term $\delta^i_t$ characterizes the deviation of the update from the accurate local objective. 
\end{assumption}
The upper bound of the one-step loss change of the global model is outlined under Assumptions \ref{l_smooth}, \ref{local_dissim}, and \ref{client_noise}, which bound the change of the global model at each global epoch. The detailed process of the proof is shown in the Appendix \ref{proof_main_theorem}.

\begin{theorem}[Bounded One-step Loss Change]\label{one_step_loss}
    Based on Assumption \ref{l_smooth}, \ref{local_dissim}, \ref{client_noise}, and Lemma \ref{bounded_dissim}, we obtain the upper bound of the loss change of the global model at each step under the noisy environment: 
    \begin{equation}
        \begin{split}
                F(w_t) \le F(w_{t-1}) + \frac{1}{2} \normtwo{\nabla F(w_{t-1})}^2 + \\ 
    \eta^2 (L+1) \normtwo{\nabla F(w_{t-1})}^2 + \eta^2 \beta^2 (L+1) + \\ 
    \eta^2 (L+1) \sum_ip_{t-1}^i \normtwo{\delta_{t-1}^i}^2
        \end{split}
    \end{equation}
\end{theorem}

\begin{remark}\label{remark_clean}
    When local clients use clean data to participate in the training, the noise term $\sum_ip_{t-1}^i \normtwo{\delta_{t-1}^i}^2$ becomes negligible. Therefore, in a clean environment, our method will have a similar performance compared to FedAvg.
\end{remark}

\begin{remark}\label{remark_noisy}
    In a noisy environment, the noisy update from a noisy client has a different or even opposite direction compared to the validation gradient, resulting in a lower influence score. Since the averaging weight $p_{t-1}^i$ is strongly based on the influence score, we can denote the relation between the average weight and the noise of the client as $p_{t-1}^i \propto \frac{1}{\delta_{t-1}^i}$. Greater noise will have a lower averaging weight, which alleviates the impact of the noise term and results in a tighter upper bound of the loss change. However, in a uniform averaging weight FedAvg, the noise term can still impact and lead to an unstable upper bound.
\end{remark}

\begin{table*}[ht]

\centering
\caption{Accuracies of Label Noise and Adversarial Samples in CIFAR-10 (\%)}
\resizebox{\textwidth}{!}{
\begin{tabular}{l|c|c|c|c|c|c|c|c|c}
\toprule
\hline

\multirow{3}*{Method} & Clean & \multicolumn{4}{c|}{Label Noise} &  \multicolumn{4}{c}{Adversarial Samples} \\
\cline{2-10}

& 0 & \multicolumn{2}{c|}{0.5} & \multicolumn{2}{c|}{0.7}  & \multicolumn{2}{c|}{0.5} & \multicolumn{2}{c}{0.7} \\

\cline{2-10}

& (0,0) & (0.5, 0.6)  & (0.7, 0.8) & (0.5, 0.6)  & (0.7, 0.8)  & (0.5, 0.6)  & (0.7, 0.8) & (0.5, 0.6)  & (0.7, 0.8) \\

\hline

\toprule
\hline

FedAvg  & $62.52 \pm 0.13$ & $54.01 \pm 0.33$ & $45.88 \pm 0.78$ & $48.22 \pm 0.35$ & $44.30 \pm 0.35$ & $59.93 \pm 0.40$ & $\mathbf{59.06 \pm 0.36}$ & $57.47 \pm 0.39$ & $55.98 \pm 0.38$ \\
FedProx & $63.11 \pm 0.35$ & $53.28 \pm 0.45$ & $45.74 \pm 0.38$ & $48.59 \pm 0.40$ & $44.29 \pm 0.23$ & $58.55 \pm 0.29$ & $58.37 \pm 0.49$ & $56.98 \pm 0.50$ & $55.50 \pm 0.27$ \\
Krum    & $47.35 \pm 1.44$ & $28.14 \pm 0.87$ & $30.21 \pm 1.22$ & $32.51 \pm 1.02$ & $26.28 \pm 1.19$ & $39.63 \pm 1.93$ & $33.29 \pm 0.39$ & $41.47 \pm 0.50$ & $38.61 \pm 0.27$ \\
AFedSV  & $63.06 \pm 0.60$ & $56.08 \pm 0.61$ & $54.57 \pm 0.51$ & $51.62 \pm 0.22$ & $49.07 \pm 0.51$ & $59.05 \pm 0.42$ & $58.93 \pm 0.32$ & $\mathbf{58.50 \pm 0.28}$ & $\mathbf{57.02 \pm 0.39}$ \\
FedIF (Ours)   & $\mathbf{63.60 \pm 0.43}$ & $\mathbf{59.12 \pm 0.36}$ & $\mathbf{55.17 \pm 0.12}$ & $\mathbf{52.99 \pm 0.75}$ & $\mathbf{51.74 \pm 0.59}$ & $\mathbf{59.98 \pm 0.42}$ & $58.90 \pm 0.21$ & $56.37 \pm 0.46$ & $55.03 \pm 0.19$ \\

\hline
\bottomrule
\end{tabular}
}
\label{main_result_cifar}
\end{table*}

\begin{table*}[ht]

\centering
\caption{Accuracies of Label Noise and Adversarial Samples in Fashion-MNIST (\%)}
\resizebox{\textwidth}{!}{
\begin{tabular}{l|c|c|c|c|c|c|c|c|c}
\toprule
\hline

\multirow{3}*{Method} & Clean & \multicolumn{4}{c|}{Label Noise} &  \multicolumn{4}{c}{Adversarial Samples} \\
\cline{2-10}

& 0 & \multicolumn{2}{c|}{0.6} & \multicolumn{2}{c|}{0.7}  & \multicolumn{2}{c|}{0.5} & \multicolumn{2}{c}{0.7} \\

\cline{2-10}

& (0,0) & (0.5, 0.6)  & (0.6, 0.7) & (0.5, 0.6)  & (0.6, 0.7)  & (0.6, 0.7)  & (0.7, 0.8) & (0.6, 0.7)  & (0.7, 0.8) \\

\hline

\toprule
\hline

FedAvg  & $89.27 \pm 0.19$ & $86.94 \pm 0.08$ & $84.01 \pm 0.41$ & $85.68 \pm 0.32$ & $84.41 \pm 0.29$ & $89.17 \pm 0.07$ & $89.08 \pm 0.09$ & $88.86 \pm 0.13$ & $88.92 \pm 0.10$ \\
FedProx & $89.24 \pm 0.18$ & $86.75 \pm 0.04$ & $83.98 \pm 0.25$ & $85.80 \pm 0.14$ & $84.33 \pm 0.33$ & $\mathbf{89.19 \pm 0.11}$ & $\mathbf{89.16 \pm 0.10}$ & $\mathbf{89.02 \pm 0.12}$ & $\mathbf{89.02 \pm 0.06}$ \\
Krum    & $84.83 \pm 0.55$ & $78.86 \pm 0.60$ & $78.36 \pm 0.86$ & $79.34 \pm 1.51$ & $77.31 \pm 1.38$ & $81.72 \pm 0.72$ & $81.48 \pm 1.01$ & $82.29 \pm 0.37$ & $82.99 \pm 0.17$ \\
AFedSV  & $\mathbf{89.64 \pm 0.11}$ & $87.31 \pm 0.18$ & $\mathbf{87.26 \pm 0.21}$ & $\mathbf{87.24 \pm 0.28}$ & $87.02 \pm 0.35$ & $89.14 \pm 0.08$ & $89.11 \pm 0.09$ & $88.90 \pm 0.17$ & $89.22 \pm 0.09$ \\
FedIF (Ours)   & $89.23 \pm 0.14$ & $\mathbf{87.95 \pm 0.13}$ & $85.93 \pm 0.23$ & $86.91 \pm 0.20$ & $\mathbf{87.02 \pm 0.22}$ & $88.93 \pm 0.13$ & $88.92 \pm 0.20$ & $88.85 \pm 0.09$ & $88.80 \pm 0.11$ \\
\bottomrule

\hline
\bottomrule
\end{tabular}
}
\label{main_result_fmnist}
\end{table*}

\section{Empirical Results}\label{experiment}

\subsection{Experimental Setting}

We evaluated our work on two image classification datasets, CIFAR-10 \cite{cifar10} and Fashion-MNIST \cite{fmnist}. CIFAR-10 consists of 50,000 training samples and 10,000 test samples, while Fashion-MNIST consists of 60,000 training samples and 10,000 test samples. Furthermore, we reserved 20\% of the test samples for the validation set. The experiments are implemented on a single Ubuntu 24.04.02 LTS machine, equipped with an NVIDIA GeForce RTX 4080 GPU, an AMD Ryzen 9 7900X 12-Core CPU, and 16GB of RAM. The source code is available at \cite{code}.

We set the total number of clients to 100 to simulate a distributed environment. Since data heterogeneity is prevalent in FL, we applied data partitioning using the Dirichlet distribution, with $\alpha_{dir}=1$ set for both datasets. To verify the robustness and efficiency of our proposed work, we first tested our method on the clean dataset as a baseline and then conducted experiments on three types of vulnerable scenarios as benchmarks:
\begin{itemize}
    \item \textbf{Label Noise}: A proportion of data labels are randomly flipped to a different class.
    \item \textbf{Gradient Noise}: The noisy client injects Gaussian noise into its local update, which is denoted by $\hat w_t^i= w_t^i + \mathcal{N}(\mu,\sigma^2)$.
    \item \textbf{Adversarial Samples} \footnote{We choose parameters for PGD as follows: $\epsilon=0.03,\alpha=0.01,iter=20,norm=L_\infty$.}: The noisy client generates the adversarial sample to participate in the local training at each global epoch by Projected Gradient Descent \cite{pgd}.
\end{itemize}

The classification accuracy and execution time are adopted to evaluate our method. For each benchmark, we denote $n\_level$ the rate of noisy clients in this system. Following the setting from \cite{fedcorr,fednoro}, we define the lower and upper bounds of the uniform distribution as $n\_ratio=(n\_lower, n\_upper)$, which are used to control the ratio of noisy samples in a noisy client for the label noise and adversarial samples benchmarks. For example, $n\_level=0.5$, $n\_ratio=(n\_lower=0.5, n\_upper=0.6)$ in the label noise scenario indicates that there are 50\% noisy clients, and each noisy client contains 50\%-60\% samples with flipped labels. We selected a different noise level for our experiments. For the gradient noise benchmark, we control the noise ratio by $\sigma$ of the Gaussian noise.

We employed CNN as the base model for the image classification task and then compared FedIF with FedAvg \cite{fedavg}, FedProx \cite{fedprox}, AFedSV \cite{AFedSV}, and Krum \cite{krum}.
For the FL hyperparameters, we selected the local batch size $B = 16$, the local epoch $E = 5$, the client fraction $C = 0.1$, and the total global epoch $T = 100$. The SGD optimizer was adopted in the local training phase, where the learning rate was set to $\eta=0.001$ and the momentum was 0.9. Specifically, the hyperparameter $\gamma$, which limits the rate of updates of the global influence value, was set to $gamma=0.3$ for CIFAR-10 and $gamma=0.4$ for Fashion-MNIST (the parameter selection process is shown in Section \ref{hyperparameter}.) 

\begin{table}[h]

\centering
\caption{Accuracies of Gradient Noise in CIFAR-10 (\%)}
\resizebox{0.48\textwidth}{!}{%
\begin{tabular}{l|c|c|c|c|c}
\toprule
\hline

\multirow{3}*{Method} & Clean & \multicolumn{4}{c}{Gradient Noise}  \\
\cline{2-6}

& 0 & \multicolumn{2}{c|}{0.5} & \multicolumn{2}{c}{0.6}  \\

\cline{2-6}

& 0 & 0.05 & 0.1 & 0.05  & 0.1  \\

\hline

\toprule
\hline

FedAvg  & $62.52 \pm 0.13$ & $56.65 \pm 0.21$ & $40.24 \pm 0.60$ & $55.87 \pm 0.53$ & $36.65 \pm 1.00$ \\
FedProx & $63.11 \pm 0.35$ & $55.92 \pm 0.40$ & $40.55 \pm 0.51$ & $54.85 \pm 0.41$ & $35.53 \pm 1.59$ \\
Krum    & $47.35 \pm 1.44$ & $47.36 \pm 0.46$ & $47.99 \pm 0.44$ & $46.85 \pm 0.47$ & $46.69 \pm 0.40$ \\
AFedSV  & $63.06 \pm 0.60$ & $\mathbf{59.61 \pm 0.42}$ & $54.34 \pm 1.02$ & $\mathbf{57.91 \pm 0.57}$ & $50.12 \pm 0.52$ \\
FedIF (Ours)   & $\mathbf{63.60 \pm 0.43}$ & $59.45 \pm 0.29$ & $\mathbf{56.09 \pm 0.25}$ & $57.59 \pm 0.40$ & $\mathbf{56.10 \pm 0.64}$ \\

\bottomrule

\hline
\bottomrule
\end{tabular}
}
\label{main_result_cifar_gradient}
\end{table}

\begin{table}[h]

\centering
\caption{Accuracies of Gradient Noise in Fashion-MNIST (\%)}
\resizebox{0.48\textwidth}{!}{%
\begin{tabular}{l|c|c|c|c|c}
\toprule
\hline

\multirow{3}*{Method} & Clean & \multicolumn{4}{c}{Gradient Noise}  \\
\cline{2-6}

& 0 & \multicolumn{2}{c|}{0.5} & \multicolumn{2}{c}{0.7}  \\

\cline{2-6}

& 0 & 0.05 & 0.1 & 0.05  & 0.1  \\

\hline

\toprule
\hline

FedAvg  & $89.27 \pm 0.19$ & $85.01 \pm 0.38$ & $79.42 \pm 0.56$ & $85.02 \pm 0.11$ & $80.00 \pm 0.44$ \\
FedProx & $89.24 \pm 0.18$ & $85.25 \pm 0.16$ & $79.51 \pm 0.55$ & $84.56 \pm 0.39$ & $79.56 \pm 0.14$ \\
Krum    & $84.83 \pm 0.55$ & $74.56 \pm 0.99$ & $73.57 \pm 0.61$ & $70.50 \pm 2.21$ & $56.28 \pm 3.31$ \\
AFedSV  & $\mathbf{89.64 \pm 0.11}$ & $\mathbf{87.42 \pm 0.20}$ & $\mathbf{85.73 \pm 0.24}$ & $\mathbf{86.90 \pm 0.15}$ & $82.99 \pm 0.22$ \\
FedIF (Ours)  & $89.23 \pm 0.14$ & $86.94 \pm 0.13$ & $83.72 \pm 0.62$ & $86.19 \pm 0.10$ & $\mathbf{83.66 \pm 0.25}$ \\

\bottomrule

\hline
\bottomrule
\end{tabular}
}
\label{main_result_fmnist_gradient}
\end{table}

\subsection{Experimental Results}
Table \ref{main_result_cifar} and Table \ref{main_result_fmnist} present the test accuracy in clean data sets, label noise, and adversarial samples in CIFAR-10 and Fashion-MNIST, respectively. Table \ref{main_result_cifar_gradient} and Table \ref{main_result_fmnist_gradient} show the accuracy of the test on a clean dataset and gradient noise in CIFAR-10 and Fashion-MNIST, respectively. We repeated our experiments five times on different random seeds and aggregated the mean value.

Our method has achieved robustness in most scenarios. FedIF and AFedSV, the other robust FL algorithm based on data valuation, only have an improvement of 1\% in CIFAR-10 and an almost equivalent performance in Fashion-MNIST compared to FedAvg. This phenomenon can be explained by Remark \ref{remark_clean}: the adaptive weight-based method only has a slight difference under a clean environment compared to the FedAvg method. However, FedIF still retains its robustness in noisy scenarios, such as label noise and gradient noise. The performances of FedIF in those noisy scenarios are close to AFedSV in most benchmarks, and even surpass this baseline method in the label noise and gradient noise in CIFAR-10. FedIF is able to achieve moderate performance in extremely noisy environments. For example, in CIFAR-10, it can achieve 51.74\% accuracy under label noise with $n\_level=0.7,n\_ratio=(0.7,0.8)$ and 56.10\% accuracy under gradient noise with $n\_level=0.6,\sigma=0.1$. However, these results for FedIF and AFedSV have merely a slight impact on the adversarial sample benchmark. In extremely noisy environments (e.g. $n\_level=0.7,n\_ratio=(0.7,0.8)$ in CIFAR-10 and $n\_level=0.7,n\_ratio=(0.7,0.8)$ in Fashion-MNIST), our method fails to provide robust performance on them. The effectiveness of FedIF relies on detecting the direction of updates, which is a single-dimensional information. However, PGD attack typically produces adversarial samples that perturb the model to make a wrong inference but preserve the similar update direction akin to that of the clean input. This similarity in update direction has the potential to mislead the valuation method, particularly when the update direction from adversarial samples aligns with the validation gradient.

Our method also demonstrates efficiency compared to AFedSV. Table \ref{execution_time} indicates the execution times (seconds) of each method, where the mean training time and the mean aggregation time are measured as metrics. Although the evaluation step in FedIF introduces extra operations and longer aggregation time compared to FedAvg, the associated computational overhead remains within reasonable limits. FedIF takes approximately 0.2 seconds during model aggregation for each global epoch, which is 450x less than AFedSV. AFedSV relies on the Monte Carlo Shapley value method to evaluate the contribution of each participant and requires multiple times of model reweighting and model inference. Instead of this time-consuming evaluation, FedIF only involves lightweight computations on the gradients of the validation set and the client updates, decreasing the computational overhead.   

\begin{table}[h]
\centering
\caption{Training Time and Aggregation Time (s)}
\begin{tabular}{l|c|c|c|c}
\toprule
\hline

\multirow{2}*{Method} & \multicolumn{2}{c|}{CIFAR-10} & \multicolumn{2}{c}{Fashion-MNIST}  \\
\cline{2-5}
& Training & Aggregation & Training & Aggregation   \\
\hline
\toprule
\hline
FedAvg  & $0.2268$ & $0.0005$ & $0.1456$ & $0.0009$  \\
FedProx & $0.2958$ & $0.0004$ & $0.2093$ & $0.0009$  \\
Krum    & $0.2326$ & $0.0015$ & $0.1456$ & $0.0019$ \\
AFedSV  & $0.2312$ & $91.7635$ & $0.1492$ & $70.8857$  \\
FedIF (Ours)  & $0.2351$ & $0.1834$ & $0.1461$ & $0.1590$ \\
\bottomrule
\hline
\bottomrule
\end{tabular}
\label{execution_time}
\end{table}

\subsection{Hyperparameter Tuning}\label{hyperparameter}

Equation (\ref{ginf}) introduces a hyperparameter $\gamma$ to control the increase rate of this global influence value. This hyperparameter is tuned according to the validation accuracy in CIFAR-10 and Fashion-MNIST under different attacks. The benchmarks for choosing $\gamma$ of FedIF are the following: clean dataset, label noise with $n\_level=0.5,n\_ratio=(0.5,0.6)$, gradient noise with $n\_level=0.5,\sigma=0.1$ and adversarial samples with $n\_level=0.5,n\_ratio=(0.5,0.6)$ for the CIFAR-10 dataset, while label noise with $n\_level=0.6,n\_ratio=(0.6,0.7)$, gradient noise with $n\_level=0.5,\sigma=0.1$ and adversarial samples with $n\_level=0.5,n\_ratio=(0.5,0.6)$ for the Fashion-MNIST dataset. We conducted the experiments five times on different random seeds and calculated the mean value of the validation accuracy for each benchmark. After considering the results of Tables \ref{cifar_gamma} and \ref{fmnist_gamma}, we select $gamma=0.3$ for CIFAR-10 and $gamma=0.4$ for Fashion-MNIST.

\begin{table}[h]

\centering
\caption{Validation accuracies with for CIFAR-10 (\%)}
\begin{tabular}{l|cccc}
\toprule

Benchmark & $\gamma=0.3$ & $\gamma=0.4$ & $\gamma=0.5$ &$\gamma=0.6$  \\
\bottomrule
\toprule

Clean & $\mathbf{63.82} $ & $61.13 $ & $62.14$ & $61.48$ \\
Label Noise  & $57.17 $ & $\mathbf{57.32} $ & $55.84$ & $56.97$  \\
Gradient Noise & $\mathbf{57.16} $ & $55.18 $ & $56.60$ & $55.70$ \\
Adv Samples & $\mathbf{57.93} $ & $57.66 $ & $57.73$ & $57.92$ \\

\bottomrule
\end{tabular}
\label{cifar_gamma}
\end{table}

\begin{table}[h]

\centering
\caption{Validation accuracies with for Fashion-MNIST (\%)}
\begin{tabular}{l|cccc}
\toprule

Benchmark & $\gamma=0.3$ & $\gamma=0.4$ & $\gamma=0.5$ &$\gamma=0.6$  \\
\bottomrule
\toprule

Clean & $\mathbf{89.45} $ & $89.41 $ & $88.99$ & $89.01$ \\
Label Noise  & $86.08 $ & $\mathbf{86.12} $ & $85.79$ & $86.11$  \\
Gradient Noise & $84.05 $ & $\mathbf{84.64} $ & $84.07$ & $85.17$ \\
Adv Samples & $88.85 $ & $\mathbf{89.28} $ & $89.33$ & $89.01$ \\
\bottomrule
\end{tabular}
\label{fmnist_gamma}
\end{table}

\subsection{Ablation Study}

To verify the efficacy of each component in our method, an additional ablation study is conducted on the CIFAR-10 data set. Specifically, we validate local weight normalization (WN), round normalization (RN), and smooth update (SU) on the same benchmark setting as Section \ref{hyperparameter} (clean data set as CL, label noise as LN, gradient noise as GN, and adversarial samples as ADV). The result of the ablation study is shown in Table \ref{ablation_result}.

The result has presented that the introduction of these components may improve the performance and robustness of our method in different dimensions. Using round normalization may offer a fair comparison of the influence value among different global rounds and make the evaluation more accurate, while a smooth update can avoid a dramatic increase in the influence value, ensuring the robustness of the model. It is noted that the local weight normalization may significantly improve the performance in the Gradient Noise scenario. Normalizing the magnitude of the local model makes the per unit of change in the client update more tangible. 

\begin{table}[ht]
\centering
\caption{Ablation Study on CIFAR-10 (\%)}
\begin{tabular}{ccc|cccc}
\toprule

WN & RN & SU & CL & LN & GN & ADV  \\
\bottomrule
\toprule

\xmark  & \cmark  & \cmark & $62.83 $ & $57.25$ & $40.46$ & $59.83$ \\
\cmark  & \xmark  & \cmark & $63.02 $ & $56.07$ & $57.75$ & $58.49$ \\
\xmark  & \xmark  & \cmark & $63.55 $ & $56.55$ & $37.46$ & $59.64$ \\
\cmark  & \cmark  & \xmark & $62.76 $ & $56.08$ & $56.43$ & $58.98$ \\
\cmark  & \cmark  & \cmark & $63.60 $ & $59.12$ & $56.09$ & $59.98$ \\
\bottomrule
\end{tabular}
\label{ablation_result}
\end{table}

\section{Conclusion}\label{conclusion}

We propose FedIF, a robust and efficient FL framework based on lightweight data valuation. Compared to Shapley-value-based methods, FedIF requires significantly less intensive calculations on the validation gradient and the updated model from clients, thereby substantially reducing computational overhead. The experimental results demonstrate that FedIF exhibits comparable performance and even surpasses AFedSV, a state-of-the-art Shapley-value-based robust FL method, in most vulnerable settings while maintaining efficiency. However, it is not effective against adversarial attacks, such as PGD, because FedIF primarily evaluates the update based on directional information. Future work will apply this method to other applications involving data valuation, such as fair incentive distribution in FL. Also, we will consider more information from the gradient to evaluate the local update.

\bibliographystyle{unsrtnat} 
\bibliography{conference_101719}

\clearpage

\appendices
\renewcommand{\theequation}{A.\arabic{equation}}
\section{Detailed Proof of Theorem \ref{one_step_loss}}\label{proof_main_theorem}
We apply the Taylor approximation to the $L$-Lipschitz smoothness function $F(w_t)$ in Assumption \ref{l_smooth} and get 
\begin{equation}
\begin{split}
    F(w_t) \le F(w_{t-1}) + \vecdot{\nabla F(w_{t-1})}{w_t-w_{t-1}} + \\
    \frac{L}{2}\normtwo{w_t-w_{t-1}}^2.
\end{split} 
\end{equation}
By AM-GM inequality, we have $\vecdot{\nabla F(w_{t-1})}{w_t-w_{t-1}} \le \frac{1}{2} \normtwo{\nabla F(w_{t-1})}^2 + \frac{1}{2} \normtwo{w_t-w_{t-1}}^2$. Therefore, we have

\begin{equation}\label{proof_main_eq}
    F(w_t) \le F(w_{t-1}) +   \frac{1}{2} \normtwo{\nabla F(w_{t-1})}^2 + \frac{L+1}{2} \normtwo{w_t-w_{t-1}}^2.
\end{equation}

\textbf{Bounding} $\normtwo{w_t-w_{t-1}}^2$:

After combining Assumption \ref{client_noise} and the rule of local update for a client $i$, we have

\begin{equation}
    w_t^i= w_{t-1} - \eta( \nabla F^i(w_{t-1}) + \delta_{t-1}^i).
\end{equation}

The aggregation function of the gloabl model follows the weighted averaging and is indicated as $w_t=\sum_i p_{t-1}^i w_t^i$, where $p_{t-1}^i$ is the averaging weight of the participant $i$ at the epoch $t-1$ and $\sum_ip_{t-1}^i=1$. According to Jensen's inequality, we get 

\begin{eqnarray}
    \normtwo{w_t-w_{t-1}}^2 &<=& \eta^2 \sum_ip_{t-1}^i \normtwo{\nabla F^i(w_{t-1}) + \delta_{t-1}^i}^2 \nonumber.
\end{eqnarray}

Applying the triangle inequality and Young's inequality to $\normtwo{\nabla F^i(w_{t-1}) + \delta_{t-1}^i}^2$, we have

\begin{equation}\label{proof_noise_ineq}
\begin{split}
    \eta^2 \sum_ip_{t-1}^i \normtwo{\nabla F^i(w_{t-1}) + \delta_{t-1}^i}^2 \le \\
    \eta^2 \sum_ip_{t-1}^i 2(\normtwo{\nabla F^i(w_{t-1})}^2 + \normtwo{\delta_{t-1}^i}^2).
\end{split}  
\end{equation}

Taking into account Lemma \ref{bounded_dissim}, the right-hand side of Equation (\ref{proof_noise_ineq}) can be bounded as

\begin{equation}
\begin{split}
    \eta^2 \sum_ip_{t-1}^i 2(\normtwo{\nabla F^i(w_{t-1})}^2 + \normtwo{\delta_{t-1}^i}^2) \le \\
    2 \eta^2 \normtwo{\nabla F(w_{t-1})}^2+ 2 \eta^2 \beta^2+  2 \eta^2  \sum_ip_{t-1}^i \normtwo{\delta_{t-1}^i}^2.
\end{split}
\end{equation}
Finish the bounding of $\normtwo{w_t-w_{t-1}}^2$. Then we have
\begin{equation}
\begin{split}
    \normtwo{w_t-w_{t-1}}^2 \le 2 \eta^2 \normtwo{\nabla F(w_{t-1})}^2+ 2 \eta^2 \beta^2+ \\
    2 \eta^2  \sum_ip_{t-1}^i \normtwo{\delta_{t-1}^i}^2.
\end{split}
\end{equation}

Combining this inequality with Equation (\ref{proof_main_eq}), Theorem \ref{one_step_loss} is proved as

\begin{equation}
    \begin{split}
    F(w_t) \le F(w_{t-1}) + \frac{1}{2} \normtwo{\nabla F(w_{t-1})}^2 + \\ 
    \eta^2 (L+1) \normtwo{\nabla F(w_{t-1})}^2 + \eta^2 \beta^2 (L+1) + \\ 
    \eta^2 (L+1) \sum_ip_{t-1}^i \normtwo{\delta_{t-1}^i}^2.
    \end{split}
\end{equation}

\bigskip\bigskip
\section{Detailed Proof of Lemma \ref{bounded_dissim}}\label{proof_bounded_dissi}

Lemma 1 is listed again here: $\sum_ip^i\normtwo{\nabla F^i(w) }^2 \le \beta^2 + \normtwo{\nabla F(w)}^2$. To prove Lemma \ref{bounded_dissim}, we start from $\sum_ip^i\normtwo{\nabla F^i(w) }^2$.

\begin{equation}
\begin{split}
 & \sum_ip^i\normtwo{\nabla F^i(w) }^2 \\
  &= \sum_ip^i\normtwo{(\nabla F^i(w) -  \nabla F(w)) +\nabla F(w) }^2  \\
     &= \sum_ip^i\normtwo{\nabla F^i(w) - \nabla F(w)}^2 + \\ 
     & 2 \sum_ip^i \vecdot{\nabla F^i(w) - \nabla F(w)}{\nabla F(w)} + \normtwo{\nabla F(w) }^2. \\
\end{split}
\end{equation}
Because the global objective is equal to the weighted averaging of local objectives, we have $\sum_ip^i \nabla F^i(w) -\nabla F(w) =0$. It cancels the second term in the above equation to 0. Therefore,

\begin{equation}
\begin{split}
  &\sum_ip^i\normtwo{\nabla F^i(w) }^2 \\
  &=\sum_ip^i\normtwo{\nabla F^i(w) - \nabla F(w)}^2  + \normtwo{\nabla F(w) }^2.
\end{split}
\end{equation}
Lemma \ref{bounded_dissim} is proved by applying Assumption \ref{local_dissim} to it: 

\begin{equation}
    \sum_ip^i\normtwo{\nabla F^i(w) }^2 \le \beta^2 + \normtwo{\nabla F(w)}^2.
\end{equation}

\end{document}